\documentclass{article} 
\usepackage{iclr2022_conference,times}

\usepackage{hyperref}
\usepackage{url}
\usepackage{graphicx}
\usepackage{amsfonts,amsmath,amssymb}
\usepackage{bbm}

\title{Convolutional autoencoders for spatially-informed ensemble post-processing}

\author{Sebastian Lerch
\\
Karlsruhe Institute of Technology\\
Karlsruhe, Germany \\
\texttt{sebastian.lerch@kit.edu} \\
\And
Kai L.\ Polsterer \\
Heidelberg Institute for Theoretical Studies\\
Heidelberg, Germany \\
\texttt{kai.polsterer@h-its.org} 
}

\iclrfinalcopy 

\begin{document}

\maketitle

\begin{abstract}
Ensemble weather predictions typically show systematic errors that have to be corrected via post-processing. 
Even state-of-the-art post-processing methods based on neural networks often solely rely on location-specific predictors that require an interpolation of the physical weather model's spatial forecast fields to the target locations.
However, potentially useful predictability information contained in large-scale spatial structures within the input fields is potentially lost in this interpolation step.
Therefore, we propose the use of convolutional autoencoders to learn compact representations of spatial input fields which can then be used to augment location-specific information as additional inputs to post-processing models. 
The benefits of including this spatial information is demonstrated in a case study of 2-m temperature forecasts at surface stations in Germany. 
\end{abstract}

\section{Introduction and Motivation}

Most weather forecasts today are based on ensemble simulations from numerical weather prediction (NWP) models, consisting of a set of deterministic forecasts that differ in initial conditions or model physics.
Despite continued improvements \citep{BauerEtAl2015}, ensemble predictions continue to exhibit systematic errors such as biases or a lack of calibration.
The process of correcting such errors to obtain more accurate and reliable forecasts is referred to as post-processing \citep{VannitsemEtAl2018book}.
Post-processing models use ensemble predictions from the NWP system as inputs, and produce a probability distribution as their output. 
Post-processing has become a standard practice in research as well as operations, and is an integral part of weather forecasting today. 
Parametric approaches from statistics where the forecast distribution takes the form of a probability distribution with parameters depending on summary statistics of the ensemble predictions of the target variable have been developed for a large variety of weather quantities.
Over the past years, much work has been spent on flexible machine learning techniques for post-processing which enable the incorporation of additional predictor variables beyond ensemble forecasts of the target variable, and have demonstrated superior forecast performance \citep{HauptEtAl2021,VannitsemEtAl2021}.
Much recent research interest has been focused on neural network (NN)-based distributional regression approaches first proposed in \citet{RaspLerch2018}, where NNs learn nonlinear relationships between arbitrary predictor variables and forecast distribution parameters in a data-driven way. 
 
All of these  post-processing methods share a common limitation: To provide predictions at individual locations (typically weather stations or grid points), they require localized ensemble forecasts,
which are obtained by interpolating the NWP ensemble members' two-dimensional forecast fields to the target locations.
However, the large-scale spatial structure and predictability information\footnote{e.g., flow-dependent error characteristics and weather regimes \citep{RodwellEtAl2018,AllenEtAl2021}} present in the physically consistent forecast fields from the ensemble simulations are lost in this interpolation step.
We propose the use of convolutional autoencoders to learn low-dimensional latent representations of the spatial forecast fields.
The learned representations are then used as additional predictors to augment a NN-based post-processing model with information about the spatial structure of relevant forecast fields.
The proposed model architecture is applied in a case study of 2-m temperature forecasts at surface stations in Germany, and compared to state-of-the-art post-processing models without spatial inputs.

The remainder of the paper is organized as follows.
Section \ref{sec:data} introduces the data.
In Section \ref{sec:methods}, we describe the autoencoders and their combination with post-processing models.
The main results are presented in Section \ref{sec:results}, and Section \ref{sec:conclusions} concludes with a discussion.
Python code with implementations of all methods is available online 
(\url{https://github.com/slerch/convae_pp}).

\section{Data}
\label{sec:data}

We use the dataset from \citet{RaspLerch2018} focusing on 2-m temperature (T2M) forecasts at surface stations in Germany at a forecast lead time of 48 h.
Forecasts from the global European Centre for Medium-Range Weather Forecasts 50-member ensemble initialized at 00 UTC every day form the basis for two types of predictor variables.
To obtain location-specific predictors, following \citet{RaspLerch2018}, we interpolate ensemble forecast of 17 meteorological variables to the observation station locations, see their Table 1 for an overview of the available variables. 
In addition, we also use the spatial forecast fields of T2M, geopotential height at 500 hPa (Z500), and the U- and V-wind at 850 hPa (U850 and V850) as a second dataset of spatial inputs.
Those variables were chosen broadly based on meteorological intuition, and are available on $0.5^\circ \times 0.5^\circ$ grid from -10E to 30E and from 30N to 70N, which roughly covers Europe and parts of the surroundings. 

Observation data of T2M for 537 weather stations in Germany are used to evaluate the forecasts.
Information about the station coordinates, altitudes and orography (altitude of the model grid point) are derived as additional input predictors for the post-processing models.
With ensemble predictions available from 3 January 2007 to 31 December 2016, we follow the setup in \citet{RaspLerch2018} and use data from 2007--2015 as training dataset, and data from 2016 as test dataset.

\section{Methods}
\label{sec:methods}

\begin{figure}[t]
	\centering
	\includegraphics[width=\textwidth]{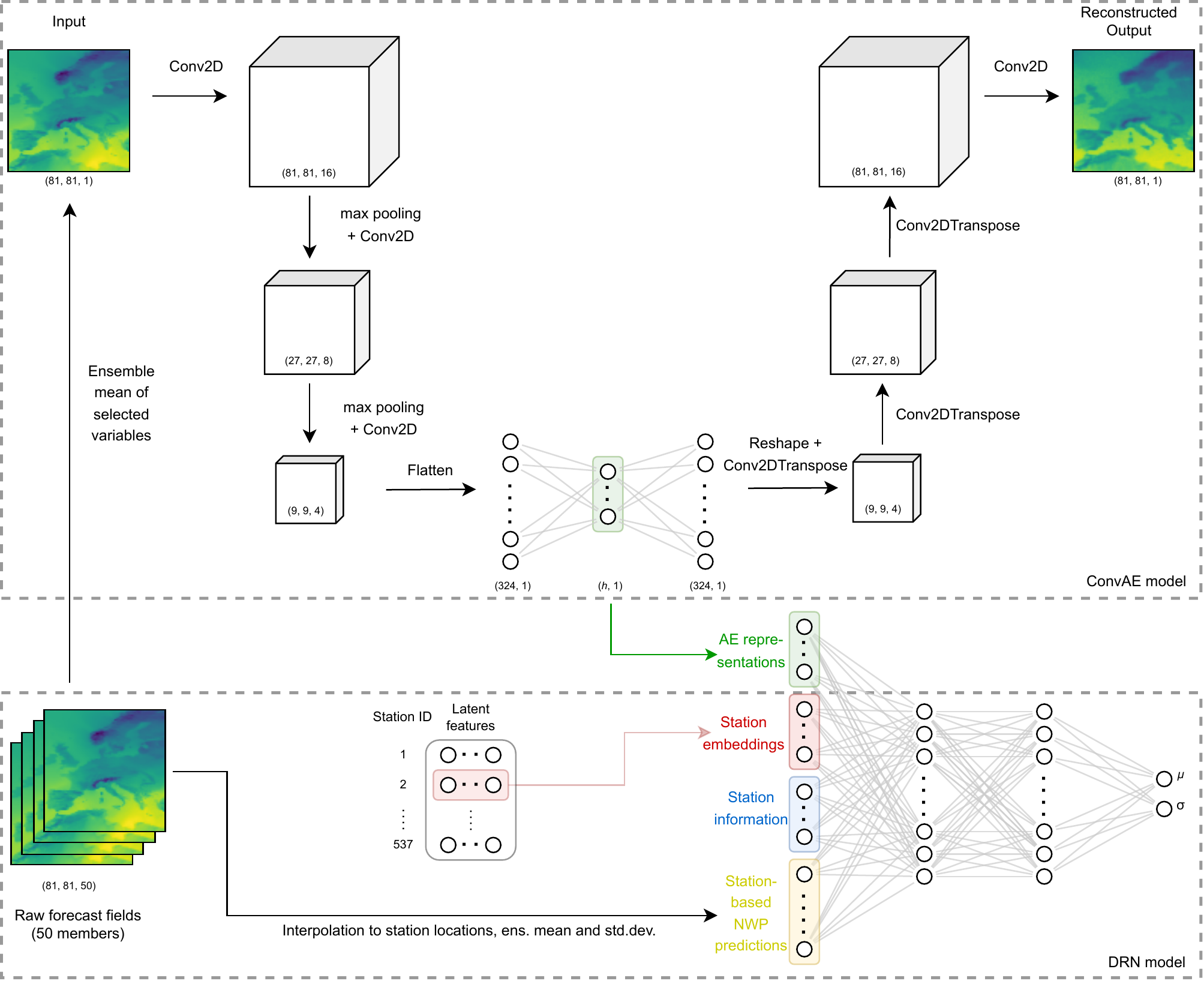}
	\caption{Schematic illustration of the DRN+ConvAE model. \label{fig:schematic}}
\end{figure}

We focus on post-processing methods within the parametric distributional regression framework proposed by \citet{GneitingEtAl2005}.
In their ensemble model output statistics (EMOS) approach, the conditional distribution of the variable of interest $y$, given ensemble predictions $\boldsymbol{X}$, is modeled by a parametric distribution, $F_{\boldsymbol{\theta}}$, with parameters $\boldsymbol{\theta} = g(\boldsymbol{X})$ depending on the ensemble predictions via a link function $g$.
The standard EMOS model for temperature utilizes ensemble forecast of temperature $\boldsymbol{X}^{\text{t2m}}$, as sole predictors and assumes a Gaussian forecast distribution $y|\boldsymbol{X}^{\text{t2m}} \sim \mathcal{N}(\mu,\sigma)$, the parameters of which are linked to the ensemble mean and standard deviation via affine functions $\mu = a + b\cdot\text{mean}(\boldsymbol{X}^{\text{t2m}})$ and $\sigma = c + d\cdot\text{sd}(\boldsymbol{X}^{\text{t2m}})$.
The model coefficients $a,b,c,d$ vary over stations (for local adaptivity), and are estimated by minimizing the mean continuous ranked probability score, 
$
\text{CRPS}(F,y) = \int_{-\infty}^{\infty} \left( F(z) - \mathbbm{1}(y \leq z)\right)^2 \text{d}z,
$
over a training set \citep{JordanEtAl2019}. 

\subsection{Neural network methods for post-processing}

A key limitation of the EMOS approach is that incorporating additional predictors beyond forecasts of the target variable is challenging since it would be necessary to specify the exact functional form of the dependencies of the distribution parameters on all input predictors.
To address this limitation, \citet{RaspLerch2018} propose to obtain the distribution parameters as the output of a NN that is able to flexibly learn nonlinear relations between arbitrary input predictors and the distribution parameters in an automated, data-driven manner.
Their distributional regression network (DRN) illustrated in the bottom part of Figure \ref{fig:schematic} is estimated as a single model jointly for all stations, using the CRPS as a custom loss function \citep{DInsantoPolsterer2018}.
Station embeddings which map the station identifiers to a vector of latent features used as additional inputs to the NN generate local adaptivity in the jointly estimated model. 
The results presented in \citet{RaspLerch2018} and subsequent research demonstrate the improvements over state-of-the-art-approaches. 

\subsection{Convolutional autoencoders as information compressor}

An autoencoder (AE) is a NN designed to learn a representation for a dataset by training the network to attempt to copy its input to its output.
Internally, a hidden layer describes an $h$-dimensional encoding used to represent the input.
Along with the encoder function, a decoder is learned that produces a reconstruction of the input data from the hidden layer.
Here, we consider separate AE models for selected weather variables (T2M, Z500, 850U, 850V) which use the spatial fields of the ensemble mean forecasts as inputs.
To account for the spatial structure of the input fields (with a size of $81\times 81$ grid points), we use 2D-convolutional layers for the encoder and corresponding transposed convolutions for the decoder, and refer to the full model as convolutional autoencoder (ConvAE), illustrated in the top part of Figure \ref{fig:schematic}.
The ConvAE model is estimated separately in a first step, using the grid point-wise mean squared error as loss function. 
Min-max normalization is applied to the individual input fields in order to guide the ConvAE models to focus on the variability across space within the forecast fields, since the relevant information on the magnitude of the predicted values is present in the station-specific, interpolated predictors for the DRN model.
Details on the model architecture and training are provided in the supplemental material.
As a reference dimensionality reduction method, we implement a principal component analysis (PCA) approach, which is widely used for different applications in the atmospheric sciences \citep[e.g.,][]{Wilks2011}.  

\subsection{Incorporating spatial inputs into NN-based post-processing}

To incorporate spatial information into the DRN model, mean forecast fields from the ensemble are used as input to the ConvAE model which was separately estimated in a first step and yields a corresponding latent space representation as output.
This latent space representation is then used as additional input to the DRN model in addition to the station-specific (interpolated) predictions, the station information and the embeddings. This combined model will be referred to as DRN+ConvAE model and is illustrated in the entirety of Figure \ref{fig:schematic}. 
We only consider spatial inputs from single predictors (T2M, Z500, 850U, 850V), T2M combined with Z500, or all of them.
To ensure comparability, the architecture and training procedure for the DRN+ConvAE models are identical to those of the DRN model without spatial inputs.
We proceed analogously for the DRN+PCA models. See Appendix \ref{sec:app-implementation} in the supplementary material for details on the model architecture and estimation.

\section{Results}
\label{sec:results}

\begin{figure}
	\centering 
	\includegraphics[height=5.45cm]{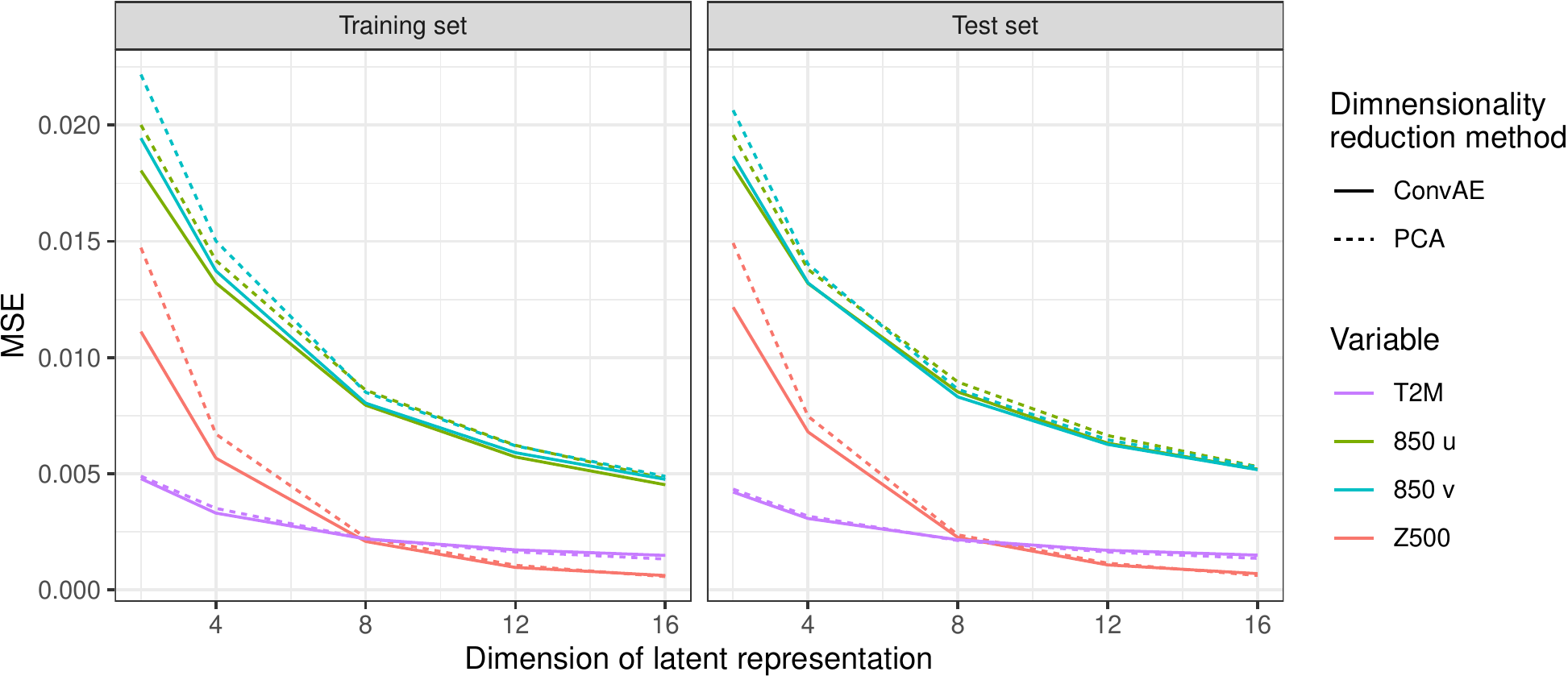}
	\caption{Grid point-wise MSE of reconstructions by the ConvAE and PCA models as function of the dimension $h$ of the latent representation for several targets on the training set (left) and the test set (right). \label{fig:dimensionality-reduction-results}}
\end{figure}

Figure \ref{fig:dimensionality-reduction-results} shows the mean squared error (MSE) of the ConvAE and PCA reconstructions. 
Since they are computed on the 0-1 scale of the normalized inputs, the errors of the different target variables are directly comparable,  
Not surprisingly, larger values of the dimension of the latent representation $h$ yield better reconstructions.
The T2M and Z500 forecast fields are generally easier to reconstruct from the low-dimensional representation than the wind fields which are characterized by small-scale structures. 
Overall, the ConvAE models show lower reconstruction errors compared to the PCA models.
The differences are most pronounced for lower-dimensional representations, and forecast fields of Z500 and the wind components, but become negligible for larger values of $h$. 
Exemplary reconstructions of the ConvAE models are shown in the supplemental material in Appendix \ref{sec:app-reconstructions}. 
Compared to the corresponding PCA reconstructions (not shown), in particular small-scale variability is notably better represented by the ConvAE models for lower values of $h$.

\begin{figure}
	\centering 
	(a)  \hfill (b) \hspace{4cm} \hphantom{1} \\
	\includegraphics[height=5.35cm]{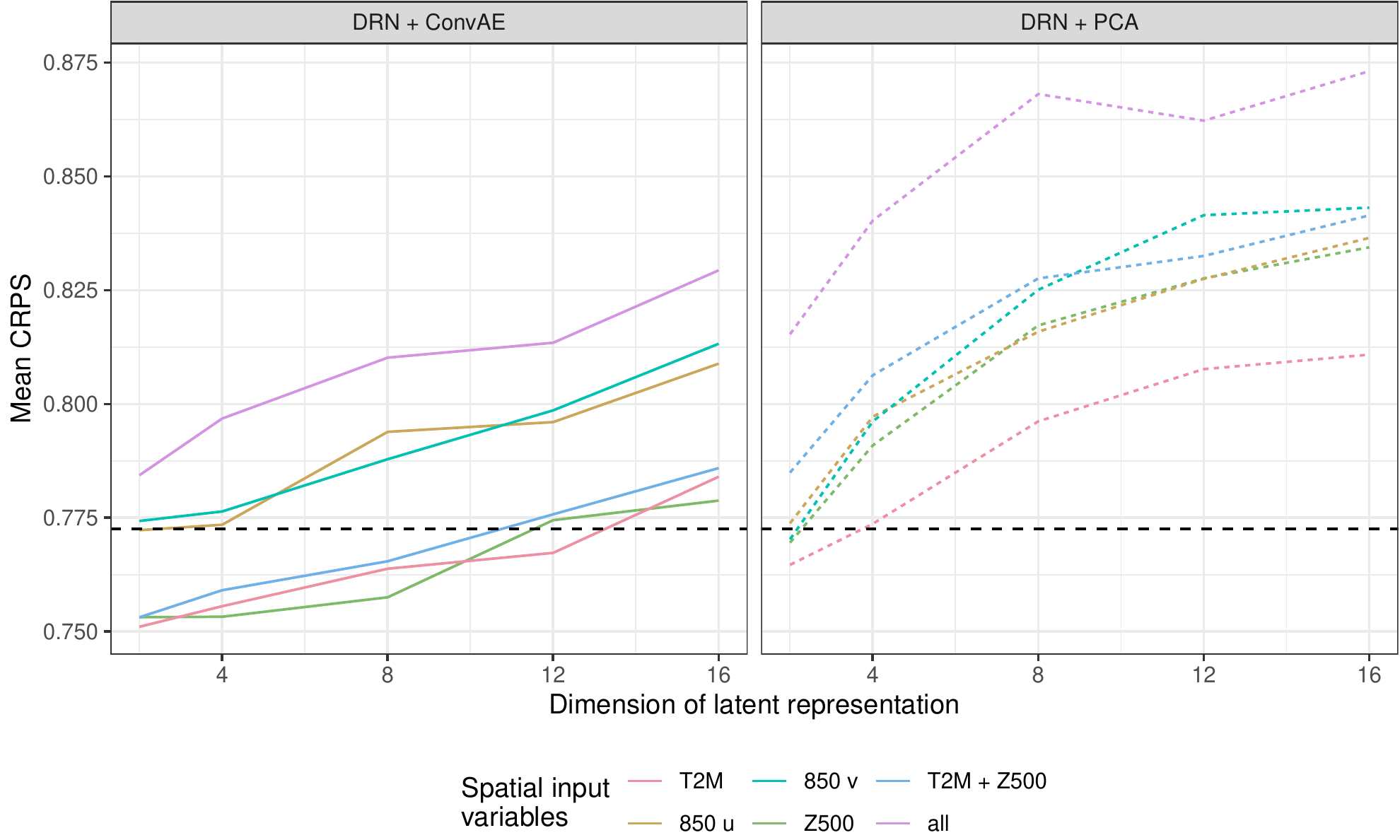}
	\hfill 
	\includegraphics[height=5.325cm]{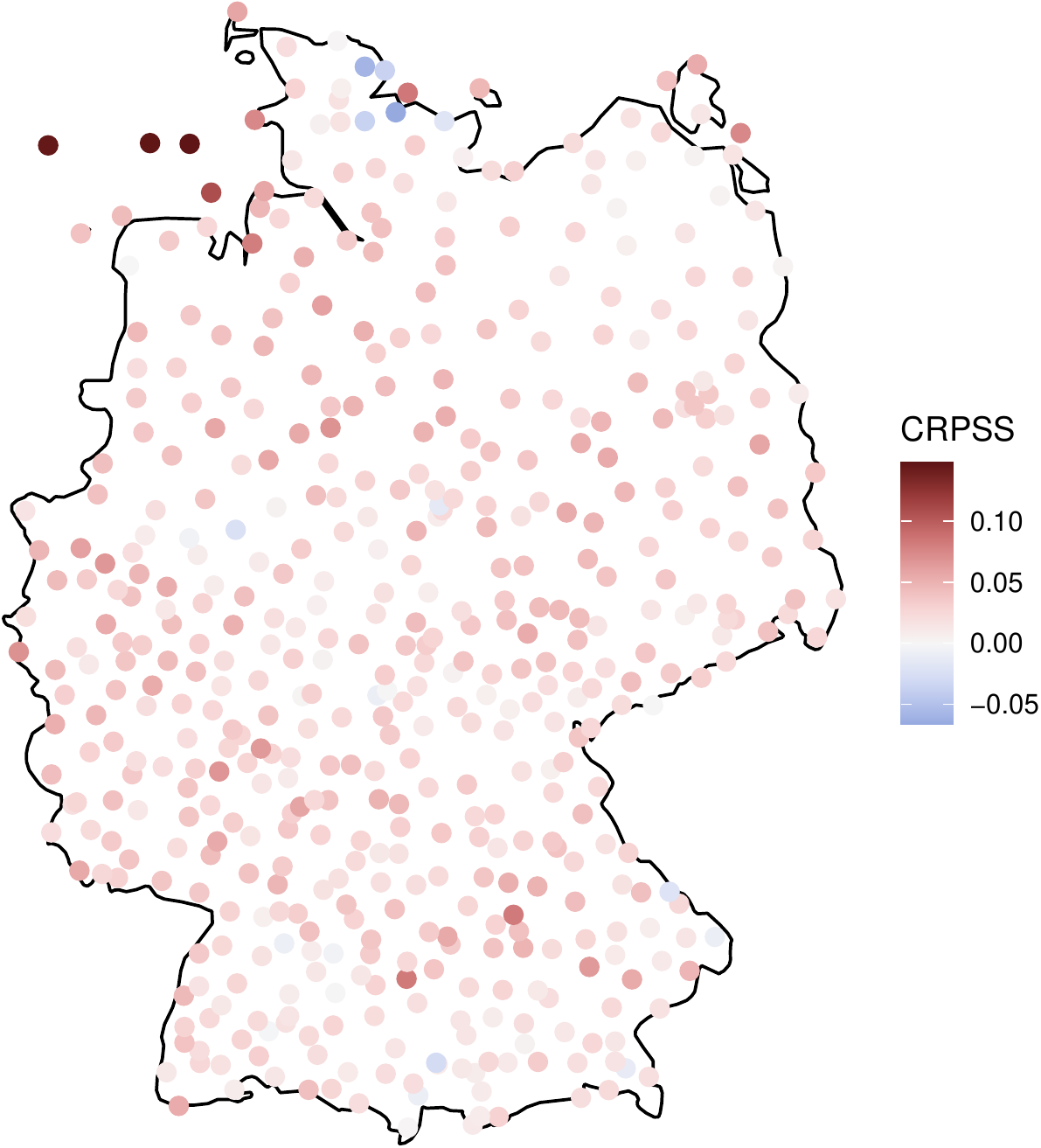}
	\caption{Mean CRPS of DRN+ConvAE (left) and DRN+PCA models (right) for different spatial inputs and values of $h$, with the CRPS of the DRN model shown as dashed black line (a); and map of station-specific improvements in terms of the CRPS of the DRN+ConvAE model with spatial T2M inputs for $h=2$, using the DRN model without spatial inputs as reference (b). \label{fig:pp-results}}
\end{figure}

Since our focus is on incorporating spatial features, we refer to \citet{RaspLerch2018}  for detailed results on the DRN model, including comparisons to other post-processing approaches.
Figure \ref{fig:pp-results}a summarizes our key results by showing the mean CRPS (averaged over all stations and dates in the test set, with lower values indicating better forecasts) of the DRN+ConvAE and DRN+PCA models as functions of the corresponding dimension of the latent representations for different sets of spatial inputs.
Improvements over DRN can be observed for DRN+ConvAE models that include spatial inputs from T2M, Z500, and their combination, for latent dimensions $h \leq 8$.
Generally, the forecast performance decreases with increasing $h$, likely since the most relevant information is already contained in lower-dimensional spatial representations.
Note that the reconstruction quality of the ConvAE and PCA models might only be of minor importance for the post-processing task of the combined model, since the added reconstruction quality may be counteracted by making it more difficult for the DRN part of the model to extract the relevant information. 
Incorporating representations from wind fields notably deteriorates the results compared to the plain DRN model. 
Directly comparing DRN+ConvAE and DRN+PCA models, it is evident that in contrast to their ConvAE counterparts, adding PCA representations only provides some minor improvements over DRN for T2M and $h=2$. 

Focusing on a comparison of the local effects of incorporating spatial inputs, Figure \ref{fig:pp-results}b shows the station-specific improvement in the mean CRPS. We here compare one DRN+ConvAE model to DRN, a CRPS skill score (CRPSS) value of 0.1 thus indicates an improvement of 10\% in the mean CRPS over DRN at that station.
Including spatial inputs in the DRN+ConvAE model results in improvements at 96\% of the stations, with Diebold-Mariano tests of equal predictive performance \citep{DieboldMariano1995} indicating that around two thirds of those improvements are statistically significant at a level of 0.05.
A comparison of the importance of the input features of the two models suggests that the DRN+ConvAE model is indeed able to extract some useful information from the ConvAE representations, see the supplemental material in Appendix \ref{sec:app-featureimportance} for details.

\section{Conclusions}
\label{sec:conclusions}

We demonstrated how information from large-scale spatial forecast fields of meteorological variables can be incorporated into post-processing models via convolutional autoencoders as information compressors.
Our post-processing models with added spatial inputs outperform state-of-the-art models that utilize station-specific predictors only. 
Regarding the design of the combined post-processing model with spatial inputs, we noted the critical need to balance the dimension of the latent representation of the spatial forecast fields:
While larger values of $h$ result in better reconstructions by the ConvAE models, they decrease the forecast performance of the corresponding post-processing model.
In addition, increasing the embedding dimension compared to the model without spatial inputs was critical to obtain improvements, likely because this allows the post-processing models to learn make  more locally adaptive use of the added spatial inputs.
While our focus was on using NN-based post-processing models, it would also be interesting to incorporate the spatial representations as additional predictors into other post-processing models, such as quantile regression forests \citep{TaillardatEtAl2016} or gradient boosting extensions of EMOS \citep{MessnerEtAl2017}.

An alternative route towards post-processing models based on spatial input data is given by the direct use of convolutional NNs (CNNs).
Forecast fields of several variables can be considered as input, and could be combined with a DRN part \citep[variants of this approach have been proposed recently in][]{ScheuererEtAl2020,VeldkampEtAl2021,ChapmanEtAl2022,LiEtAl2022}.
A comparison to the methods proposed here is not straightforward since those approaches typically use gridded observations as target variables, but provides an interesting starting point for future research. 

Finally, the ConvAE models proposed here raise several interesting methodological questions.
While we only used mean fields as inputs, it would be interesting to consider probabilistic encoders in order to make better use of the ensemble structure of the ensemble members' forecast fields, which can be interpreted as samples from a spatial probability distribution.
Further, a meteorological analysis of the learned representations of the spatial forecast fields, for example considering links to weather regimes, would not only be interesting from a meteorological perspective, but might also allow for better incorporating physical information and constraints into the forecasting models.

\subsubsection*{Acknowledgments}
Sebastian Lerch gratefully acknowledges support by the Vector Stiftung through the Young Investigator Group ``Artificial Intelligence for Probabilistic Weather Forecasting.'' K.L. Polsterer gratefully acknowledges the support of the Klaus Tschira Foundation. We thank Antonio D'Insanto, Tilmann Gneiting, and Peter Knippertz for helpful comments and discussions.

\bibliography{aepp_bib}
\bibliographystyle{iclr2022_conference}

\newpage
\appendix
\renewcommand{\figurename}{Supplementary Figure}
\setcounter{figure}{0}

\section{Appendix: Supplemental material}

\subsection{Model specification details and sensitivity experiments}
\label{sec:app-implementation}

\subsubsection*{ConvAE models}

The final architecture of the ConvAE model shown in Figure \ref{fig:schematic} was chosen based on preliminary experiments on the training dataset. The encoder part consists of a sequence of convolutional (with 16, 8 and 4 filters) and max-pooling layers. The 2D-convolution layers use $3\times 3$ kernels with a stride of 1, zero-padding and a ReLU activation. The number of filters and the kernel size were chosen to balance computational costs and representation quality, but the results were generally fairly robust to changes in these parameters. 
The max-pooling layers use a window size of $3\times 3$ and a stride of 1. 
In the central dense encoding layer, we apply a linear activation function and ReLU activations in the neighboring dense layer. Following the standard practice in the extant literature \citep[e.g.,][]{RonnebergerEtAl2015} the decoder part of the ConvAE model is based on 2D-transposed convolution layers, here with 4, 8 and 16 filters, a decoder kernel of size $9\times 9$ and ReLU activations. The final output is obtained via a 2D-convolution layer with sigmoid activation.
The model is trained using the Adam optimizer with a learning rate of 0.001 and a batch size of 32. To prevent overfitting, we set the maximum number of epochs to 100 and apply early stopping with a patience of 10. Thereby, data from 2007--2014 is used for training, and data from 2015 as a validation dataset.

\subsubsection*{DRN and DRN+ConvAE/PCA models}

\begin{figure}[h]
	\includegraphics[width=\textwidth]{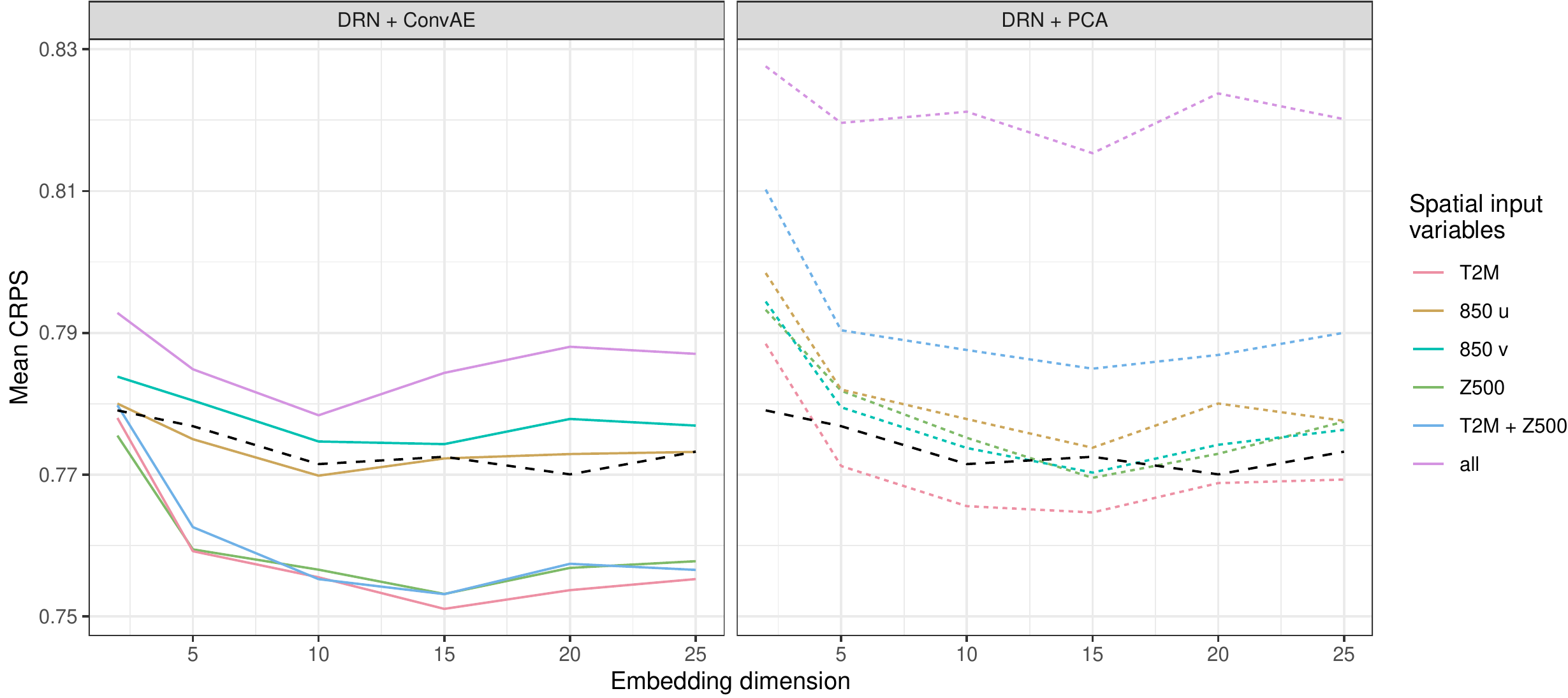}
	\caption{Mean CRPS over the test set as a function of the embedding dimension for the DRN+ConvAE and DRN+PCA models, with different types of spatial inputs and $h=2$. The black dashed line indicates the mean CRPS of the DRN model without spatial inputs. \label{fig:embdim}}
\end{figure}

The DRN and the DRN+ConvAE/PCA models share a common architecture to enable a fair comparison and a direct investigation of the effect of including spatial inputs. 
All models use two hidden layers with 100 nodes and ReLU activations. 
While including a second hidden layer deviates from the choices in \citet{RaspLerch2018}, we found that this has only a negligible effect on the performance of the DRN model, but does improve the models with spatial inputs. 
The models are trained using the Adam optimizer with a learning rate of 0.002 for a maximum of 100 epochs, and early stopping with a patience of 10 is applied to prevent overfitting. 
We produce an ensemble of NN models by repeating the model estimation 10 times, and aggregate the predictions by averaging the distribution parameters.
Data from 2007--2014 is used for training, and data from 2015 for validation purposes.
An important tuning parameter is the dimension of the station embeddings. While \citet{RaspLerch2018} use two-dimensional embeddings, subsequent research demonstrated the usefulness of choosing larger values \citep[e.g.,][]{Bremnes2020,SchulzLerch2022}. We chose an embedding dimension of 15 for all models. Supplementary Figure \ref{fig:embdim} shows the mean CRPS as functions of the embedding dimension and indicates that for the models with added spatial inputs, choosing larger than 2 improves the forecast performance, wheres the effects on the DRN model are relatively minor. 

\clearpage 

\subsection{Exemplary ConvAE reconstructions}
\label{sec:app-reconstructions}

\begin{figure}[h]
	\centering
	\includegraphics[width=0.89\textwidth]{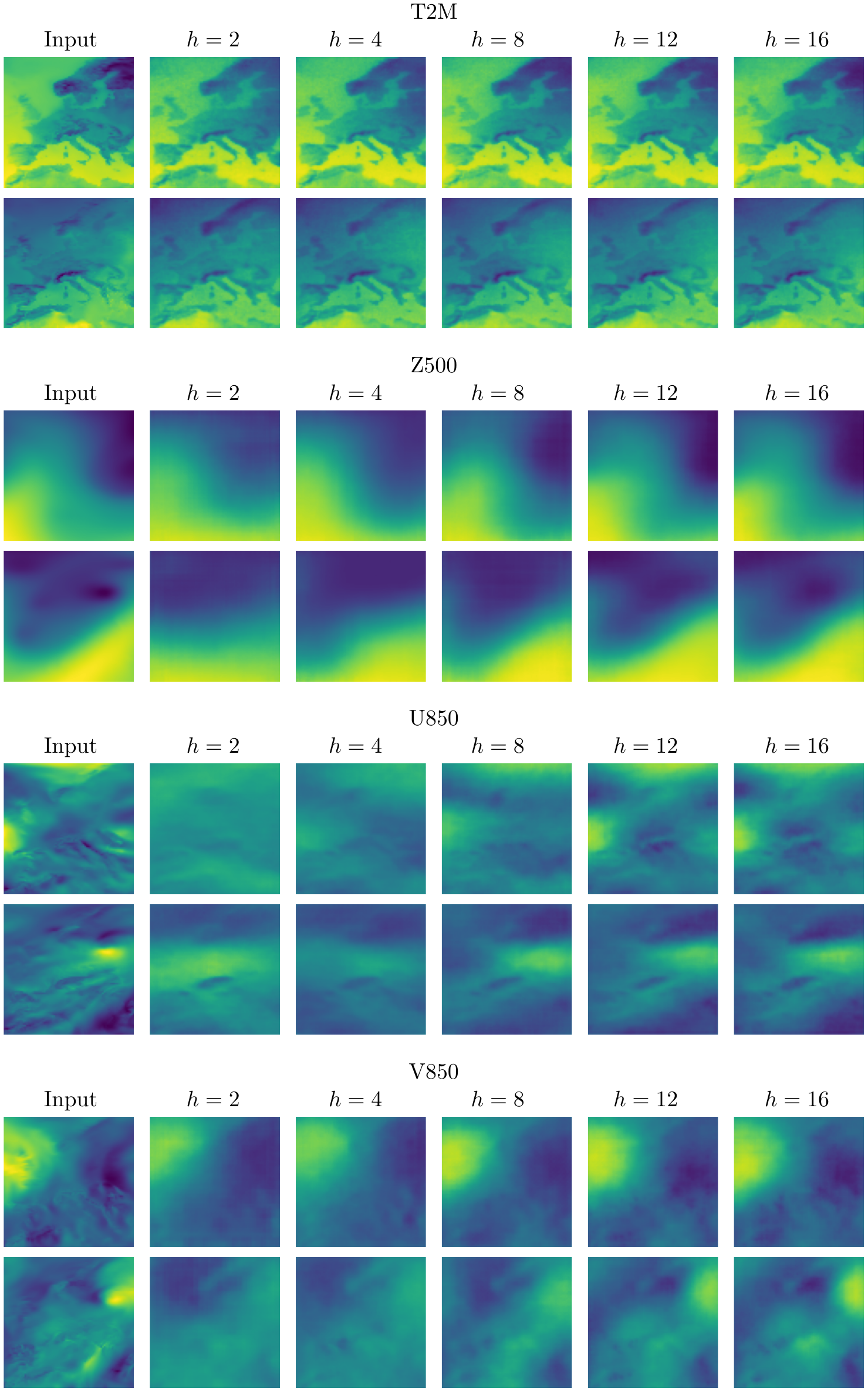}
	\caption{Exemplary ConvAE reconstructions of randomly selected examples from the test dataset for different values of the encoding dimension $h$.}
\end{figure}

\clearpage 

\subsection{Feature importance}
\label{sec:app-featureimportance}

\begin{figure}[h]
	\centering
	\includegraphics[width=\textwidth]{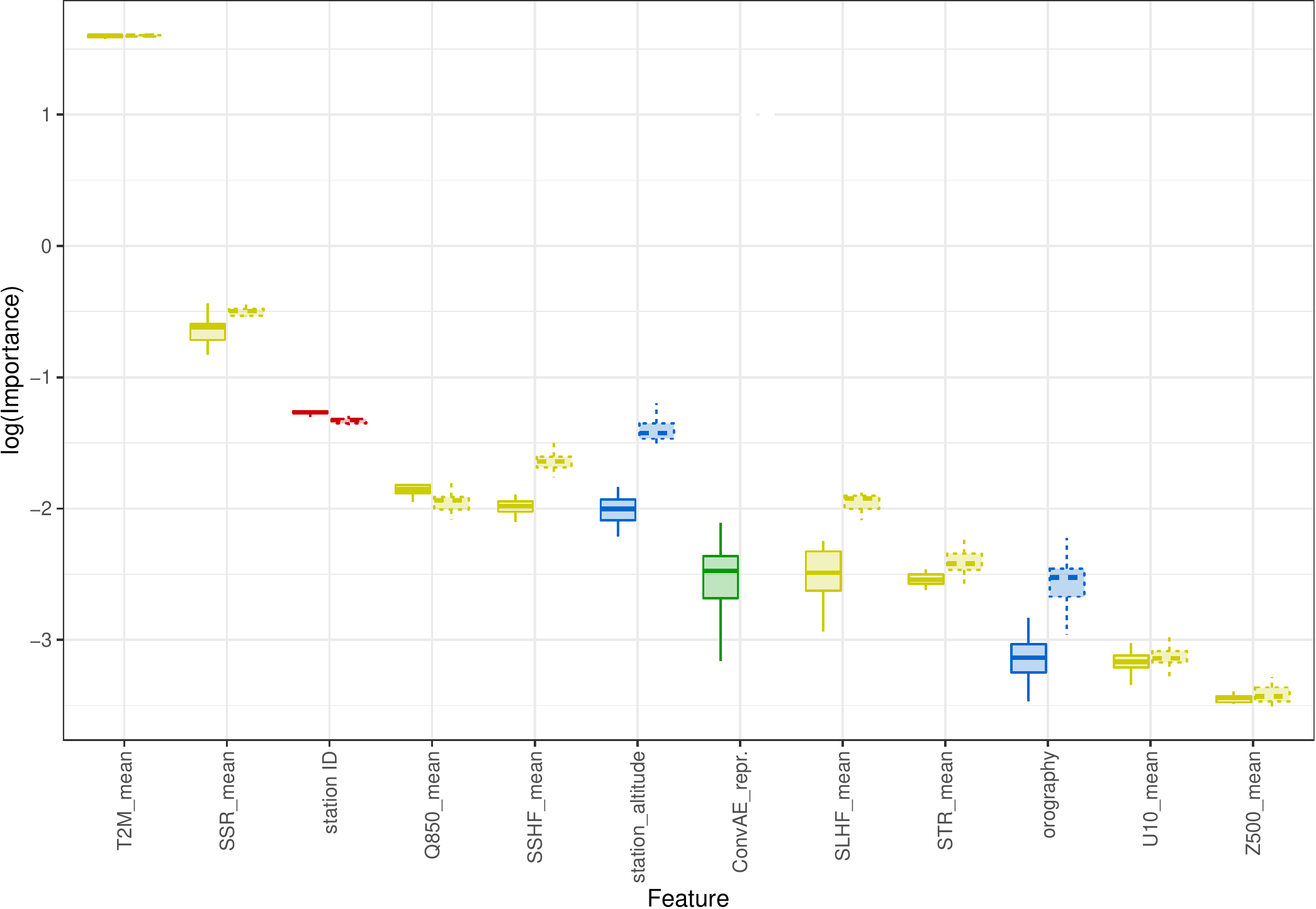} \\
	\includegraphics[width=\textwidth]{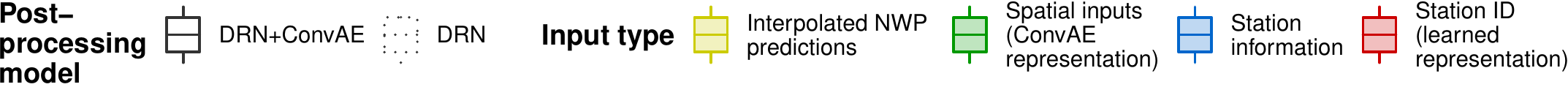} 
	\caption{Permutation-based feature importances of the 12 most importance predictors of the DRN+ConvAE (with T2M inputs and $h=2$) and DRN model shown on a logarithmic scale. The input features are colored by type according to the illustration in Figure \ref{fig:schematic} in the main text, see \citet{RaspLerch2018} for abbreviations of the interpolated NWP variables. The boxplots indicate the variability of the importances across the 10 repetitions of the model fitting procedure. \label{fig:fimp}}
\end{figure}

Supplementary Figure \ref{fig:fimp} shows the permutation-based feature importance of the 12 most important predictors of the DRN+ConvAE and the DRN model. To compute feature importances, we follow \citet{RaspLerch2018} and \citet{SchulzLerch2022}, and measure the decrease in terms of the CRPS in the test set when randomly permuting a single input feature, using the mean CRPS of the respective model based on unpermuted input features as reference.

Overall, the rankings among the most important features are relatively consistent among the two models, but a decreased importance can be observed for the DRN+ConvAE model for most of the features compared to the DRN model, most notably for the station altitude and orography. The feature importance of the ConvAE representations of the spatial T2M inputs ranks seventh on average, but shows a notably larger variability over repetitions of the model fitting procedure compared to the other inputs.

\end{document}